\documentclass[11pt,letterpaper]{article}
\usepackage[T1]{fontenc}
\usepackage[margin=1in]{geometry}
\usepackage{newtxtext}
\usepackage{amsmath,amssymb}
\usepackage{microtype}
\usepackage{graphicx,booktabs,float}
\usepackage[authoryear,round]{natbib}
\setcitestyle{citesep={;},aysep={,},yysep={;}}
\usepackage{titlesec}
\usepackage[font=small,labelfont=bf,labelsep=period,skip=7pt]{caption}
\usepackage[section]{placeins}
\usepackage[hidelinks]{hyperref}
\usepackage{url}

\usepackage{amsmath,amsfonts,bm}

\def\eqref#1{equation~\ref{#1}}

\def\1{\bm{1}}

\DeclareMathAlphabet{\mathsfit}{\encodingdefault}{\sfdefault}{m}{sl}
\SetMathAlphabet{\mathsfit}{bold}{\encodingdefault}{\sfdefault}{bx}{n}

\newcommand{\tableone}{
\begin{tabular}{lccccc}
\toprule
model & seed 0 & seed 1 & seed 2 & mean & above threshold? \\
\midrule
matrix fast-weight & 0.9995 & 1.0000 & 0.9990 & 0.9995 & every seed \\
flat-vector recurrence & 0.0322 & 0.0327 & 0.0369 & 0.0339 & never \\
transformer (uncapped) & 0.0271 & 0.0293 & 0.0286 & 0.0283 & never \\
\midrule
\multicolumn{6}{l}{$\Delta$ vs flat-vector: mean 0.9656, 95\% paired CI (0.9582, 0.9729)} \\
\multicolumn{6}{l}{$\Delta$ vs transformer: mean 0.9712, 95\% paired CI (0.9685, 0.9738) (descriptive)} \\
\bottomrule
\end{tabular}}
\newcommand{\tablecapped}{
\begin{tabular}{lccc}
\toprule
model / KV budget & $T{=}454$ & $T{=}902$ & $T{=}1798$ \\
\midrule
matrix, fixed 32{,}768 B & 0.9989 & 0.9989 & 0.9989 \\
transformer, uncapped & 0.0293 & 0.0314 & 0.0339 \\
transformer, $M{=}32$ cap & 0.0300 & 0.0251 & 0.0256 \\
transformer, $M{=}16$ cap & 0.0280 & 0.0240 & 0.0251 \\
transformer, $M{=}8$ cap & 0.0257 & 0.0264 & 0.0250 \\
transformer, $M{=}4$ cap & 0.0251 & 0.0277 & 0.0251 \\
transformer, $M{=}2$ cap & 0.0251 & 0.0286 & 0.0257 \\
\midrule
\multicolumn{4}{l}{per-$M$ paired-gap CI lower bounds at $T{=}902$: all $\geq 0.9586$} \\
\bottomrule
\end{tabular}}
\newcommand{\tabletaps}{
\begin{tabular}{llcccc}
\toprule
model & linear readout & cos.\ mean & rf@0.9 & shuffled cos.\ & gap \\
\midrule
matrix & $S_1 q$ (shallow query) & 0.165 & 0.000 & 0.105 & +0.060 \\
matrix & $S_0 q$ (bind-path query) & 0.118 & 0.000 & 0.112 & +0.006 \\
matrix & $S_1 q$ (true query) & 0.167 & 0.000 & 0.104 & +0.063 \\
matrix & pre-LM-head hidden & 0.894 & 0.674 & 0.094 & +0.800 \\
\midrule
flat-vector & $S_1 q$ (shallow query) & 0.117 & 0.000 & 0.112 & +0.005 \\
flat-vector & $S_0 q$ (bind-path query) & 0.113 & 0.000 & 0.111 & +0.002 \\
flat-vector & $S_1 q$ (true query) & 0.118 & 0.000 & 0.113 & +0.005 \\
flat-vector & pre-LM-head hidden & 0.119 & 0.000 & 0.112 & +0.006 \\
\bottomrule
\end{tabular}}
\newcommand{\tabletasktwo}{
\begin{tabular}{lccccccccc}
\toprule
model & s0 & s1 & s2 & s3 & s4 & s5 & s6 & s7 & s8 \\
\midrule
matrix & 0.040 & 0.029 & \textbf{0.334} & 0.036 & 0.031 & \textbf{0.479} & 0.028 & \textbf{0.391} & 0.041 \\
flat-vector & 0.034 & 0.028 & 0.037 & 0.031 & 0.031 & 0.032 & 0.034 & 0.031 & 0.035 \\
\bottomrule
\end{tabular}}
\newcommand{\tablekstress}{
\begin{tabular}{lccc}
\toprule
 & matrix fast-weight & flat-vector & transformer \\
\midrule
$K{=}48$ ($K/d{=}0.75$) accuracy & 0.0189 & 0.0195 & 0.0218 \\
\multicolumn{4}{l}{chance 0.0208; exploratory threshold 0.0625; no model above threshold} \\
\bottomrule
\end{tabular}}
\newcommand{\nparamscontender}{14{,}049{,}408}
\newcommand{\nparamsablation}{14{,}048{,}384}
\newcommand{\nparamstransformer}{14{,}440{,}448}

\titleformat{\section}{\large\bfseries}{\thesection}{0.65em}{}
\titleformat{\subsection}{\normalsize\bfseries}{\thesubsection}{0.65em}{}
\titlespacing*{\section}{0pt}{15pt plus 2pt minus 2pt}{7pt}
\titlespacing*{\subsection}{0pt}{10pt plus 2pt minus 1pt}{5pt}
\renewenvironment{abstract}{%
  \par\vspace{4pt}\begin{center}\bfseries Abstract\end{center}
  \small\noindent\ignorespaces
}{\par\vspace{9pt}}

\begin{document}
\begin{center}
{\fontsize{18}{21}\selectfont\bfseries Constant-Memory Recall\par}
\vspace{5pt}
{\fontsize{13}{16}\selectfont
Learned Associations in a Fixed Matrix State\par}
\vspace{13pt}
{\normalsize Samuel Larson\par}
{\small Pebble ML\quad\texttt{samlarson@pebbleml.com}\par}
\end{center}
\vspace{5pt}

\begin{abstract}
Fixed-size recurrent memory limits storage growth during inference, but
successful recall depends on the task and training. We study a small
DeltaNet variant with fixed token-specific key biases, trained to remember 32 new key--value pairings per sequence.
With 32\,KiB of recurrent matrix state, it achieves 99.95\% mean accuracy across
three training seeds when choosing among the sequence's values.
Recall remains near perfect when filler extends the pre-query context
to 1{,}798 tokens without adding pairings. Zeroing the first memory block removes
this recall. 
An exploratory 48-pair test remains near chance after one quarter of
the primary training budget and does not locate a capacity limit.
Parameter-matched vector and Transformer baselines remain near chance,
including the Transformer after additional training searches. This
unresolved baseline failure prevents a memory-efficiency comparison.
\end{abstract}

\section{Introduction}

A Transformer's key--value cache grows with context length
\citep{vaswani2017attention}; a fast-weight model instead updates a
fixed-size memory as tokens arrive \citep{schlag2021linear}. This can
bound the memory used during inference, but retaining a fixed set of
facts for longer is different from storing more facts or following a
chain of associations.

Learned associative recall with fast weights is established
\citep{schmidhuber1992learning, ba2016using, schlag2021linear}, including
successful multi-query recall with DeltaNet \citep{yang2024parallelizing}.
We examine retrieval and filler retention in one small DeltaNet variant,
with an exploratory check at higher load. State interventions
and linear probes test how the learned model uses its memory.

The comparisons use models of approximately 14 million parameters,
trained from scratch on the same synthetic episodes. The matrix model
learns the primary task; the vector and Transformer baselines do not.
The vector control also has less memory and a different write operation.
Because the Transformer fails even with its full cache, the comparison
cannot establish a memory-efficiency advantage over attention. We report
the trained model's behavior and the unsuccessful controls, without
attributing their difference to matrix structure alone.

\section{Experimental setup}
\label{sec:setup}

\subsection{Recall task and scoring}

Each episode samples $K=32$ distinct entity tokens from a GPT-2 token
pool \citep{radford2019language} and links them in a random cycle. Each
entity is a key whose value is its successor, so keys and values share
the same entity set. Each pairing occupies seven tokens, giving a
224-token binding sequence. Entities and the cycle are redrawn for every
episode. A six-token query names a key and asks for its successor.

\textbf{Scoring.} At the answer position, we take the highest-scoring
LM-head output among the episode's $K$ values. This
\emph{episode-restricted accuracy} has chance level $1/K=0.03125$;
the pre-registered recall threshold is $3/K=0.09375$. It measures discrete
retrieval, not full-vocabulary generation, state rank, or exact recovery
of continuous value vectors.

\subsection{Models and training setup}

\textbf{Matrix fast-weight model.} A two-block decoder-style model holds a
$64\times64$ state per block, with the DeltaNet update
\citep{schlag2021linear, yang2024parallelizing}
\[
S_t=S_{t-1}(I-\beta_t k_t k_t^{\top})+\beta_t v_t k_t^{\top},
\qquad o_t=S_tq_t.
\]
Keys blend their learned projections with fixed random token-specific
vectors: $k_t=\operatorname{normalize}(0.42\,k_t^{\rm raw}+0.58\,b_{x_t})$.
The two fp32 matrix states occupy 32{,}768 bytes, independent of context
length. This counts the fast-weight matrices, excluding model weights,
the fixed bias tables, convolution history, and temporary activations.
The two fixed tables occupy an additional 24.54\,MiB and are excluded
from the parameter counts below.

\textbf{Vector recurrence.} The embedding table, output head, and
feed-forward blocks are unchanged. The recurrence is
$s_t=s_{t-1}\odot g_t+v_t$, with readout $o_t=s_t\odot q_t$ and state
width 64. Mixer width is adjusted to match total parameters. Its state
occupies 512 bytes, $64\times$ less than the matrix model, and its update
lacks a key-dependent outer product. Thus this control changes both
memory size and write operation; no memory-matched, key-conditioned
vector control was tested. 

\textbf{Transformer.} A two-layer GPT-2-style causal decoder uses the same
tokenizer and width. Evaluation uses either its full KV cache or
attention-sink-plus-FIFO eviction \citep{xiao2023efficient}, with total
cache limits of $M\times32{,}768$ bytes for
$M\in\{1,2,4,8,16,32\}$. 

The matrix, vector, and Transformer models have \nparamscontender{},
\nparamsablation{}, and \nparamstransformer{} trainable parameters, respectively,
a maximum difference of 2.8\%. The primary runs train for 20{,}000 steps
at learning rate $3\times10^{-4}$ with the same optimizer, batch schedule,
and episode stream, with eight query continuations per training episode.
The objective sums next-token cross-entropy,
answer-token cross-entropy, and a target-vector cosine loss weighted by
2.0. Paired seeds receive identical episodes.

\subsection{Analysis}

Three training seeds are evaluated on the same 4{,}096 held-out queries
over 128 episodes; query keys are sampled with replacement. The pre-registered primary comparison requires matrix
recall above $3/K$ and a paired-seed 95\% $t$-interval on the
matrix-minus-vector accuracy difference whose lower bound exceeds 0.30.
These intervals measure training-seed variation on a fixed query set
($\mathrm{df}=2$), not independent variation across 4{,}096 samples.
Other comparisons are descriptive, with no multiple-comparison correction.
An additional test with hidden hop counts is documented separately in
Appendix~\ref{app:ambiguous}. Cache limits are interpreted as a
memory-efficiency test only if the full-cache Transformer first learns
recall.

\section{Results}
\label{sec:results}
\subsection{Single-step recall}
\label{sec:recall}

\begin{table}[htbp]
\centering
\small
\tableone
\caption{Recall of 32 pairings after 20{,}000 training steps, by training
seed. Accuracy is restricted to the episode's values; chance is 0.031
and the recall threshold is 0.094. The matrix model learns the task in
all three seeds; both parameter-matched baselines remain near chance.}
\label{tab:main}
\end{table}

The matrix model reaches mean accuracy 0.9995. Its paired advantage over
the vector recurrence is 0.9656 (95\% interval 0.9582--0.9729), exceeding
the primary comparison's 0.30 margin. The descriptive difference against
the Transformer is 0.9712 (0.9685--0.9738). 
These are outcomes under a shared training recipe; neither baseline
learns recall despite falling training loss
(Appendix~\ref{app:traincurve}).

A separate Transformer search tested eleven configurations with three
seeds each, varying capacity, learning rate, and training length up to
44.6 million parameters and 60{,}000 steps. No evaluated seed exceeded
the recall threshold; configuration means ranged from 0.0283 to 0.0346.
Appendix~\ref{app:baseline} gives the full search design. The cause of
this failure remains unresolved, and the search does not exhaust
possible training recipes.

\FloatBarrier
\subsection{Recall over longer sequences}
\label{sec:horizon}

\begin{figure}[htbp]
\centering
\includegraphics[width=\linewidth]{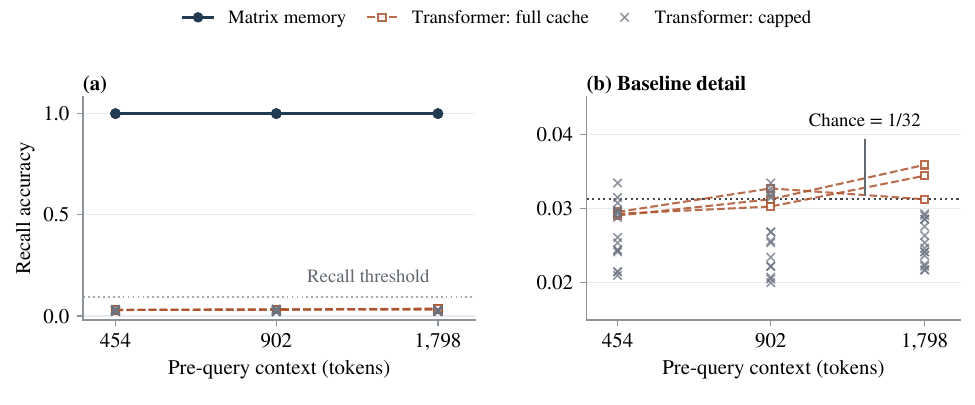}
\caption{Recall after extending the pre-query context with filler, keeping 32
pairings fixed. Panel (a) shows the full accuracy range; (b) enlarges the
baseline scores. The matrix model retains accuracy $\geq0.998$ with
32\,KiB of matrix state across three seeds. Full-cache (dashed) and capped-cache
(crosses, $M\in\{2,4,8,16,32\}$) Transformers remain near chance.
Reference lines mark the recall threshold and chance.
The protocol excludes $M=1$ from this plot; its scores are reported in
Table~\ref{tab:capped}.}
\label{fig:horizon}
\end{figure}

We evaluate frozen checkpoints with pre-query contexts of 454, 902, and
1{,}798 tokens, appending repeated buffer tokens after the 224-token
binding sequence.
The six query tokens follow this context.
No new pairings are added. The matrix model scores 0.9995, 0.9983, and
0.9990 by seed at every tested length, using the same 32{,}768-byte state.
Transformer scores remain near chance with either a full or capped cache
(Figure~\ref{fig:horizon}; Appendix Table~\ref{tab:capped}). These tests
show retention through filler, not resistance to new interfering
associations or an advantage over a successful attention baseline.

\FloatBarrier
\subsection{State interventions and readout probes}
\label{sec:mechanism}

\begin{figure}[htbp]
\centering
\includegraphics[width=0.88\linewidth]{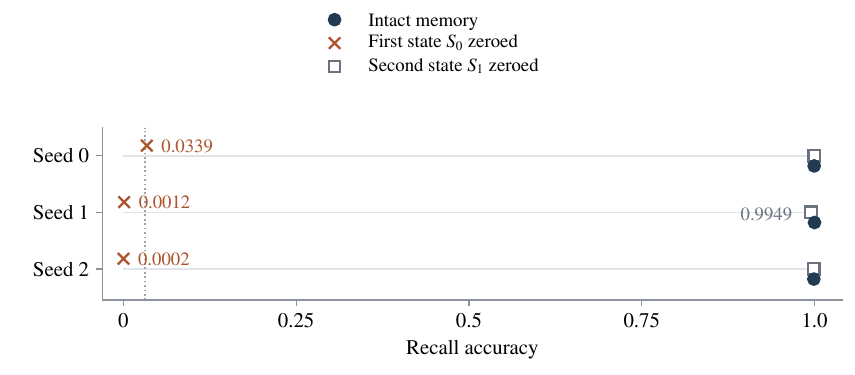}
\caption{Zeroing a memory block before the query, using 4{,}096 fixed
queries per seed. Removing $S_0$ reduces matrix-model accuracy below
the 0.094 recall threshold; removing $S_1$ leaves it nearly unchanged.
Chance is $1/32$. Seed markers are offset vertically for readability.}
\label{fig:szero}
\end{figure}

Zeroing the first block's state $S_0$ after the binding sequence reduces
accuracy to 0.0339, 0.0012, and 0.0002 across seeds. Zeroing $S_1$ gives
0.9995, 0.9949, and 0.9990, with 21 fewer correct answers out of 4{,}096 in seed 1.
The query is evaluated from the saved matrix states and new query
tokens, without replaying the binding sequence.

Ridge probes fitted on 24{,}576 fresh examples fail to recover target
value vectors from any tested raw-state/query readout above cosine 0.9.
For the matrix model's hidden state before the LM head, the probe reaches
mean cosine 0.894 and threshold pass fraction 0.674 on a separate
20{,}000-step seed-0 calibration checkpoint (Appendix Table~\ref{tab:taps}). 
Regularization was selected on the same evaluation set, so these are
descriptive results. Failure of these probes establishes neither that
all linear readouts fail nor that storage itself is nonlinear.

\FloatBarrier
\subsection{Exploratory higher-load test}
\label{sec:scope}

\begin{table}[htbp]
\centering
\small
\tablekstress
\caption{Recall with 48 pairings: one 5{,}000-step run per model, versus
20{,}000 steps for the primary 32-pair comparison. All scores are near
chance (0.021). The shorter training budget prevents attributing the
failure to memory load alone.}
\label{tab:k48}
\end{table}

The 48-pair check uses the same state width but only one quarter of the
primary training budget. Its near-chance scores
(Table~\ref{tab:k48}) therefore establish neither a capacity boundary nor
failure at a matched training budget.

\section{Related work}
\label{sec:related}

Fast-weight memory has long been used for temporary associations
\citep{schmidhuber1992learning, ba2016using}. \citet{schlag2021linear}
connect it to linear attention \citep{katharopoulos2020transformers},
introduce the delta-rule update, and test retrieval, overwriting, and
memory load. DeltaNet subsequently achieved perfect accuracy on its
hardest tested multi-query recall setting \citep{yang2024parallelizing}.
Gated DeltaNet distinguishes repeated filler from interfering natural
text \citep{yang2024gated}. These establish the mechanism and recall
capability studied here.

Zoology \citep{arora2023zoology} finds attention stronger than the recurrent
alternatives it tests; \citet{okpekpe2025revisiting} show that optimization
and model dimensions affect recall comparisons. Our terminal query and
restricted candidate set differ from multi-query recall. Comparisons
that match both parameters and recurrent-state size
\citep{hatamizadeh2026gated} also control a confound present in our vector
baseline. Vector representations can support key-dependent binding
\citep{plate1995holographic, smolensky1990tensor}; our elementwise control
does not test those alternatives.

Cache-reduction methods \citep{xiao2023efficient, zhang2023h2o} ask how
much cache preserves a working model's behavior. Our unsuccessful
full-cache baseline prevents that comparison. Capacity results for
associations stored in trained parameters \citep{nichani2024understanding}
address a different setting from fresh pairings stored in recurrent
state.

\section{Limitations}
\label{sec:limitations}

These experiments test one small fast-weight architecture trained from
scratch on synthetic episodes. They do not establish performance on
pretrained models, natural text, or deployment-length contexts. Longer
sequences contain filler rather than new bindings. Only three training
seeds and one fixed query set support the primary result. The higher-load
check has a shorter training budget, and a separate hidden-hop test cannot
measure multi-hop ability (Appendix~\ref{app:ambiguous}). The failed Transformer and smaller-memory vector
control prevent an architectural-efficiency claim. Independently tuned
baselines and a memory-matched vector control would be needed to explain
the observed difference.

\section{Conclusion}

A small DeltaNet variant learns to retrieve 32 new pairings from a fixed
32\,KiB matrix state and retains them through the tested filler sequences.
Its first memory block is necessary for recall. The experiments do not establish how this result extends to higher
loads, queries with specified hop counts, or natural text.
These results document this operating point and its limits;
they establish neither a general capacity law nor a memory-efficiency
advantage over attention.

\clearpage
\begingroup
\small\raggedright
\setlength{\bibsep}{5pt}
\bibliographystyle{colm2026_conference}
\bibliography{refs}
\endgroup

\clearpage
\appendix
\section{Recall across cache sizes}
\label{app:capped}

\begin{table}[H]
\centering
\small
\tablecapped
\caption{Mean recall across three seeds at pre-query context lengths
$T$. The matrix state is fixed at 32{,}768 bytes; Transformer cache caps
are $M$ times that size. Chance is 0.031 and the recall threshold is
0.094. The protocol excluded the $M=1$ eight-token cap because it is below the
registered floor of 13 tokens (query plus one binding clause). Its
descriptive scores, 0.021--0.030, are also near chance.}
\label{tab:capped}
\end{table}

\section{Linear readout results}
\label{app:taps}

\begin{table}[H]
\centering
\small
\tabletaps
\caption{Ridge-probe recovery of target value vectors on separate 20{,}000-step seed-0 calibration checkpoints. Probes fit 24{,}576 examples and evaluate 4{,}096 queries;
regularization was selected on that evaluation set. ``rf@0.9'' is the
fraction above cosine 0.9; ``gap'' subtracts a shuffled-fit control.
Only the matrix model's pre-LM-head hidden state supports substantial
recovery. These are descriptive probe results.}
\label{tab:taps}
\end{table}
\vspace{8pt}

``Shallow query'' applies the second block's query pathway to query-token
embeddings. ``Bind-path query'' uses the first block's pathway.
``True query'' is the second block's query during full continuation.
Across the three primary training seeds, pre-LM-head probe pass fractions
are 0.686, 0.771, and 0.951. 

\section{Training loss}
\label{app:traincurve}

\begin{figure}[H]
\centering
\includegraphics[width=0.88\linewidth]{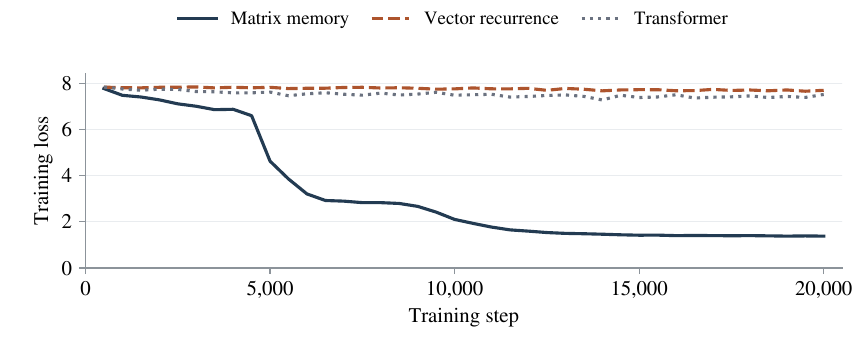}
\caption{Training loss for the primary task, seed 0. After 20{,}000 steps,
loss is 1.38 for the matrix model, 7.69 for the vector recurrence, and
7.51 for the Transformer. The baselines' decreasing loss does not produce
recall. These runs use the shared settings of Section~\ref{sec:setup}.}
\label{fig:traincurve}
\end{figure}

\section{Transformer search and reproducibility}
\label{app:baseline}
\label{app:repro}

The Transformer search considers two layers at width 256
(14{,}440{,}448 parameters), four at width 256 (16{,}014{,}336), and six at
width 512 with eight heads (44{,}613{,}632). Learning rates are $10^{-3}$
and $3\times10^{-4}$; training lengths are 20{,}000 and 60{,}000 steps.
Eleven configurations have three seeds each: 27 new runs and six reused
runs. The six-layer, $3\times10^{-4}$, 60{,}000-step configuration was
not run under the registered stopping rule. No evaluated seed exceeds
the 0.09375 recall threshold. 

The figure script reads archived evaluation artifacts and verifies their
checksums. The accompanying evidence map records input paths and maps
reported measurements to their source artifacts.

\section{Diagnostic with an unspecified hop count}
\label{app:ambiguous}

An additional experiment selects targets one or two steps along the
episode's cycle without telling the model which. The archived code uses
hop count to select the target but omits it from the input. These
ambiguous-target results (Table~\ref{tab:task2}) cannot establish either
multi-hop capability or a failure to learn it.

\begin{table}[H]
\centering
\small
\tabletasktwo
\caption{Recorded accuracy under ambiguous hop supervision, nine seeds
per model. Bold values exceed the original 0.094 threshold. The numbers
are retained for documentation; crossing that threshold is not evidence
of following a requested hop count, which was absent from the input.}
\label{tab:task2}
\end{table}

\end{document}